%% file: main.tex
\documentclass{article}

\usepackage[preprint]{neurips_2026}
\makeatletter\renewcommand{\@noticestring}{}\makeatother % no footer notice
\usepackage[utf8]{inputenc}
\usepackage[T1]{fontenc}
\usepackage{hyperref}
\hypersetup{hidelinks}
\usepackage{url}
\usepackage{booktabs}
\usepackage{longtable}
\usepackage{enumitem}
\usepackage{amsfonts}
\usepackage{amsmath,amssymb}
\usepackage{microtype}
\usepackage{xcolor}
\usepackage{graphicx}
\usepackage{caption}
\usepackage{tikz}
\usepackage{wrapfig}
\usepackage{xcolor}
\usetikzlibrary{positioning,arrows.meta}

\title{Many Pixels, One Composition: Self-Supervised Spectra for Specimen-Level Prediction}

\author{%
  Eleftheria Tetoula-Tsonga \\
  Institute of Communication and Computer Systems\\
  Athens, Greece \\
  \And
  George Arvanitakis \\
  Geonova \\
  Athens, Greece \\
  \And
  Theodoros Giannakas \\
  Institute of Communication and Computer Systems\\
  Athens, Greece \\
}

\begin{document}
\maketitle

\input{sections/00_abstract_TG}
\input{sections/01_introduction_TG}
\input{sections/04_method_TG}
\input{sections/06_results_TG}
\input{sections/07_conclusion}

\bibliographystyle{plainnat}
\bibliography{references}

\end{document}

%% file: sections/00_abstract_TG.tex
\begin{abstract}
Can hyperspectral (HS) imaging be used to predict the elemental composition of physical specimens found on the Earth's surface? We approach this problem as a supervised regression task, using an open dataset of specimens, each comprising a HS cube and elemental characterization. 
In particular, we train an encoder in a self-supervised fashion that operates on pixel-level; then freeze the encoder and train a lightweight attention-based regression head in a supervised manner, by pairing many representations of a given specimen with its corresponding elemental label.
In test specimens, our approach significantly outperforms traditional spectral matching and unmixing baselines. It also outperforms supervised learning models based on the mean spectrum (of similar model capacity), while full model fine-tuning achieves the strongest performance, at substantially higher model capacity.
\end{abstract}

%% file: sections/01_introduction_TG.tex
\section{Introduction}

% Elemental composition determines the economic value of a rock specimen; however, it is a time-consuming, as it requires laboratory assays. 
Inferring the elemental composition of physical specimens found on the Earth’s surface typically relies on techniques, such as X-ray-based sensing, which can be slow and difficult to scale. To circumvent this, non-destructive estimation from hyperspectral short wave infrared (SWIR) imagery could directly exploit surface spectra: a single hyperspectral scan is fast, requires no sample preparation, and can be repeated on the same specimen.

% To resolve this problem, standard workflows from signal processing literature match pixels against spectral libraries such as USGS, either through hard assignment or spectral unmixing \citep{kokaly2017usgs,heinz2001fcls,iordache2011sparse}, then convert mineral abundances into elemental composition via stoichiometric rules.
A conventional first step is to match pixel spectra against reference libraries such as the USGS, using either hard assignment or spectral unmixing~\citep{kokaly2017usgs,heinz2001fcls,iordache2011sparse}. These methods are unsupervised by nature. However, an advantage they have over a supervised learning counterpart is that they can rely on a prior library of well-selected spectral signatures, without requiring paired HS observations and composition labels from target specimens.

% they have is that they have a large library which they can use as look-up, and hence, unlike  relying on such a huge of reference spectra, their advantage is that they can cover accuracy depends on library coverage and conversion fidelity, rather than on the size of any paired composition dataset.

In the hyperspectral ML domain, learned encoders have been applied for tasks such as land-cover classification and scene reconstruction on \emph{satellite or airborne} imagery \citep{hong2024spectralgpt,wang2025hypersigma,zhu2023spectralmae}, under acquisition conditions (spatial context, illumination, sensor band configuration) that differ substantially from close-range specimen scans. These models, nonetheless, operate on a different pixel scale (in the order of many meters) and solve problems with different supervision signals.

% Adapting these models to specimen-level composition prediction would require the same specimen-level supervision used in this work, and no established recipe exists for doing so; they are consequently not evaluated here. General-purpose multiple-instance learning frameworks \citep{zaheer2017deep,ilse2018attention} have likewise not previously been applied to quantitative composition estimation from spectra.
A recently released dataset, ``Minerals in the Wild''~\citep{tetoula2026minerals}, is a collection of specimens, each accompanied by a hyperspectral cube and a corresponding XRF measurement providing the elemental composition of the specimen.
% A recently released dataset provides attempts to hyperspectral cubes (HSC) paired with one specimen-level XRF label (\emph{i.e.}, a vector including the elemental concentration of the scanned specimen)~\citep{tetoula2026minerals}.
%
We treat a specimen as a bag of unordered pixels (spectra) paired with its XRF supervision, essentially casting the problem as a multiple instance regression. To this end, we pretrain a spectrum/pixel-level encoder, freeze it, and sample many pixel embeddings per specimen. Finally, a permutation-invariant gated attention \citep{zaheer2017deep,ilse2018attention} pools these into a specimen-level representation from which a regression head computes the elemental composition of the whole specimen.

% This pairing creates a granularity mismatch: each cube contains thousands of pixel spectra, whereas supervision is available once per specimen. We formulate specimen-level composition prediction as multiple instance regression over pixel spectra: rather than predicting from a single averaged spectrum or requiring pixel-level labels, we pretrain a compact spectrum-level encoder on close-range cubes, freeze it, and sample many pixel embeddings $z_i$ per specimen; permutation-invariant gated attention \citep{zaheer2017deep,ilse2018attention} aggregates these into a fixed-dimensional representation 
% $Z$, from which a compact head learns $f(Z)=y$ under specimen-level supervision alone. This confines supervised training to a lightweight head and accommodates variable-sized specimens without explicit mineral identification.

% We evaluate two complementary questions: whether direct XRF-supervised prediction is a viable alternative to library-based pipelines, which are not trained on paired specimen composition data and therefore serve as conventional references rather than equal-information baselines; and whether, at matched downstream trainable capacity, the frozen set-based model uses identical XRF supervision more effectively than a mean-spectrum model. The results show that at matched trainable capacity the frozen set-based model outperforms mean-spectrum alternatives, while full encoder fine-tuning attains the highest accuracy at substantially greater trainable capacity.

%% file: sections/04_method_TG.tex
\section{Predicting elemental composition from hyperspectral cubes}

\subsection{Problem statement}

The dataset consists of 1{,}132 rock specimens with 2{,}587{,}554 valid pixels, in total; each specimen has a different pixel count.
Pixels are arranged as an image for each specimen. 
Each pixel's spectrum lies in $\mathbb{R}^{D}$, where $D=273$ are the light reflectance values in the SWIR range (1000--2500\,nm); while each specimen $s$ has an associated XRF label $y_s \in \mathbb{R}^{M}$, a vector of concentrations across the $M=48$ elements measured by the sensor. We use a fixed 905/113/114 train/validation/test specimen split, with all pixels of a given specimen kept in the same partition to avoid pixel-level leakage.

\paragraph{Task.} Let specimen $s$ be represented as a set of $n_s$ unordered valid pixels, \emph{i.e.}, $X_s = \{x_{s,1}, \dots, x_{s,n_s}\}$, where $x_{s,p}\in \mathbb{R}^{D}$ is the spectrum of pixel $p$ and specimen $s$. We seek to learn a function that predicts the specimen composition vector directly from $X_s$, using only specimen-level supervision. That is, to learn:
\begin{align}
f : X_s \in \mathbb{R}^{n_s \times D} \to \hat{y}_s \in \mathbb{R}^{M}
\end{align}

\paragraph{Metrics.} To evaluate the performance of $f$, we compute the $L_2$ distance and cosine similarity between predicted $\hat{y}_s$ and ground-truth $y$ composition vectors. The former captures whether predictions are close in absolute magnitude. The latter quantifies whether a prediction has the same nonzero elements as the true measurement.

\subsection{Solution: embeddings-based prediction}

\begin{figure}[t]
\centering
\includegraphics[width=\linewidth]{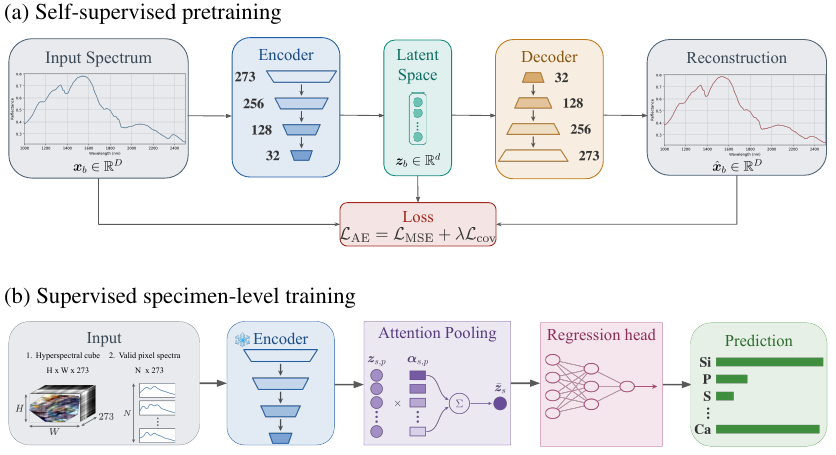}
\caption{Two-stage model: self-supervised pixel encoding followed by gated-attention pooling and specimen-level regression.}
\label{fig:arch}
\end{figure}

\paragraph{Self-supervision on pixels.} To exploit the large number of pixels, we leverage self-supervised learning. Each specimen is represented by $N=512$ randomly selected pixels. Each training batch contains $K=8$ specimens. Let a training batch contain $B = K\cdot N$ spectra from the training set, that is, $X = [x_1,\ldots,x_B]^\top \in \mathbb{R}^{B\times D}$, where each $x_b\in\mathbb{R}^{D}$ is a pixel spectrum. An encoder maps each spectrum to a $d$-dimensional representation,
$z_b=e(x_b)\in\mathbb{R}^{d}$, and a decoder reconstructs the input spectrum as $\hat{x}_b$. We collect the representations into $Z=[z_1,\ldots,z_B]^\top\in\mathbb{R}^{B\times d}$. We train the autoencoder using a reconstruction loss together with a covariance regularizer:
\begin{equation}
\mathcal{L}_{\mathrm{AE}}=
\frac{1}{BD}
\sum_{b=1}^{B}
\left\lVert x_b-\hat{x}_b\right\rVert_2^2
+
\frac{\lambda}{d^2-d}
\sum_{i\neq j}\bigl[C(Z)\bigr]_{ij}^2,
\end{equation}
where $[C(Z)]_{ij}$ is the entry $(i,j)$ of the covariance matrix, evaluated on the $B$ representations. The first term measures reconstruction error across the $B$ pixels and their spectral bands in the batch, while the second penalizes correlations between latent dimensions, encouraging a non-redundant representation~\citep{bardes2022vicreg}. After pretraining, the decoder is discarded and the encoder is frozen for the subsequent specimen-level prediction task. The self-supervised stage of the proposed approach is illustrated in Figure~\ref{fig:arch}(a).

% \subsection{Specimen-level supervised regression}

\paragraph{Supervised learning: from specimen representations to specimen label.} Each specimen contains multiple heterogeneous pixel spectra but only one XRF composition vector. For a given $s$, each pixel is projected independently by the frozen encoder, yielding representations $z_{s,p}\in\mathbb{R}^{d}$ (where we chose $d=32$). We then aggregate these  using the gated attention pooling operator of \citet{ilse2018attention}. The attention weights and the pooled representation are computed, respectively, as follows:
\begin{equation}
\alpha_{s,p} =
\frac{
\exp\!\left(
w^\top
\left[
\tanh(Vz_{s,p})
\odot
\sigma(Uz_{s,p})
\right]
\right)}
{
\sum_q
\exp\!\left(
w^\top
\left[
\tanh(Vz_{s,q})
\odot
\sigma(Uz_{s,q})
\right]
\right)
},
\qquad
\bar{z}_s = \sum_p \alpha_{s,p} z_{s,p},
\end{equation}
where $p,q$ run over the $N$ pixels of specimen $s$; and $V,U\in\mathbb{R}^{64\times32}$ and $w\in\mathbb{R}^{64}$ are the learnable matrices. This pooling operation allows the model to assign different weights to individual spectra rather than averaging them uniformly. 
A $32\to128\to48$ MLP maps $\bar z_s$ to the specimen-level composition $\hat y_s$, and the trainable components are optimized using as loss $\mathcal{L}_{\text{H}}=\operatorname{smooth}{L_1}(\hat y_s-y_s)$, applied element-wise. The subsequent supervised specimen-level regression stage is illustrated in Figure~\ref{fig:arch}(b).

% A $32\to128\to48$ MLP maps $\bar z_s$ to the specimen-level composition $\hat{y}_s$; the blocks after the encoder are optimized using smooth-$L_1$ loss between $\hat{y}_s$ and $y_s$.

%% file: sections/06_results_TG.tex
\section{Performance evaluation: results and discussion}

\begin{wrapfigure}[18]{r}{0.5\textwidth}
\vspace{-0.5\intextsep}
\centering
\captionsetup{font=scriptsize}
\includegraphics[width=\linewidth]{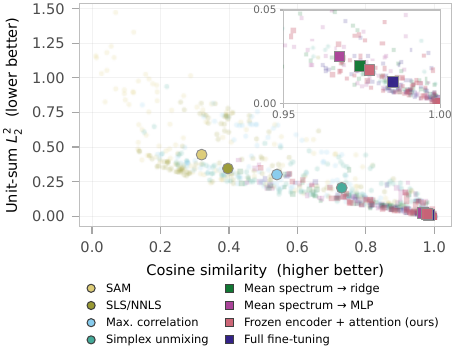}
\caption{Specimen-level performance. XRF-supervised models cluster in the high-cosine, low-error region, whereas the spectral matching and unmixing baselines show broader errors.}
\label{fig:scatter}
\end{wrapfigure}

\begin{table}
\vspace{-\intextsep}
\centering
\caption{Performance on 114 test specimens. Results for neural-network-based methods are mean $\pm$ standard deviation over five seeds. $L_2^2$ is computed after unit-sum normalization.}
\label{tab:main}
% \footnotesize
% \captionsetup{font=footnotesize}
\small
\setlength{\tabcolsep}{4pt}
\begin{tabular}{@{}lccc@{}}
\toprule
Method & Trainable Params & Cosine $\uparrow$ & $L_2^2$ $\downarrow$ \\
\midrule
SAM & -- & 0.4013 & 0.5251 \\
SLS/NNLS & -- & 0.4601 & 0.3850 \\
Max. correlation & -- & 0.5639 & 0.3344 \\
Simplex unmixing & -- & 0.6820 & 0.2267 \\
\midrule
Mean spectrum + ridge & 13.1k & 0.9041 & 0.0568 \\
Mean spectrum + MLP & 14.5k &
$0.9011{\pm}0.0021$ & $0.0654{\pm}0.0021$ \\
\textbf{Frozen enc.\ + attn. (ours)} & \textbf{14.8k} &
$\mathbf{0.9282{\pm}0.0042}$ & $\mathbf{0.0479{\pm}0.0021}$ \\
Full fine-tuning & 122.7k &
$0.9415{\pm}0.0011$ & $0.0394{\pm}0.0011$ \\
\bottomrule
\end{tabular}
\end{table}

\paragraph{Comparison.} We consider two groups of baselines. \emph{First} consists of four unsupervised methods: SAM, Max. correlation, and two methods based on convex optimization, found in~\cite{tetoula2026minerals}. All four perform spectral matching or unmixing without using the XRF measurements. \emph{Second}, it is supervised learning methods. We consider two models trained on specimen-level XRF measurements, using the mean spectrum of each specimen as input, as well as a model that fine-tunes the spectral encoder jointly with the specimen-level predictor. We evaluated them on the fixed 114 specimen test set using cosine similarity and unit-sum $L_2^2$.
% We reimplemented the four mineral-first methods described in \emph{Minerals in the Wild} and These use no XRF supervision and therefore serve as conventional references rather than equal-information baselines. We additionally compare against XRF-supervised mean-spectrum models at matched downstream trainable capacity, isolating the effect of the set-based representation under identical label access.
The unsupervised methods exhibit broad specimen-level variation, whereas the XRF-supervised models are concentrated near the high-cosine, low-error region (Fig.~\ref{fig:scatter}). Our model also outperforms both mean-spectrum models, reducing unit-sum $L_2^2$ from 0.0568 to 0.0479 relative to the ridge baseline (Table~\ref{tab:main}). Replacing ridge with a similarly sized nonlinear MLP does not improve the mean-spectrum baseline, suggesting that the gain does not arise simply from additional nonlinearity or parameter count. This result is consistent with the hypothesis that retaining the pixel spectra and pooling them as a set preserves specimen-level information lost through global averaging. Full encoder fine-tuning achieves the highest accuracy, but the frozen model remains close while requiring approximately 88\% fewer trainable parameters (Table~\ref{tab:main}). This suggests that the frozen representation already contains much of the spectral information required for elemental prediction.

\begin{wrapfigure}[14]{r}{0.5\textwidth}
\vspace{-0.3\intextsep}
\centering
\captionsetup{font=scriptsize}
\includegraphics[width=0.72\linewidth]{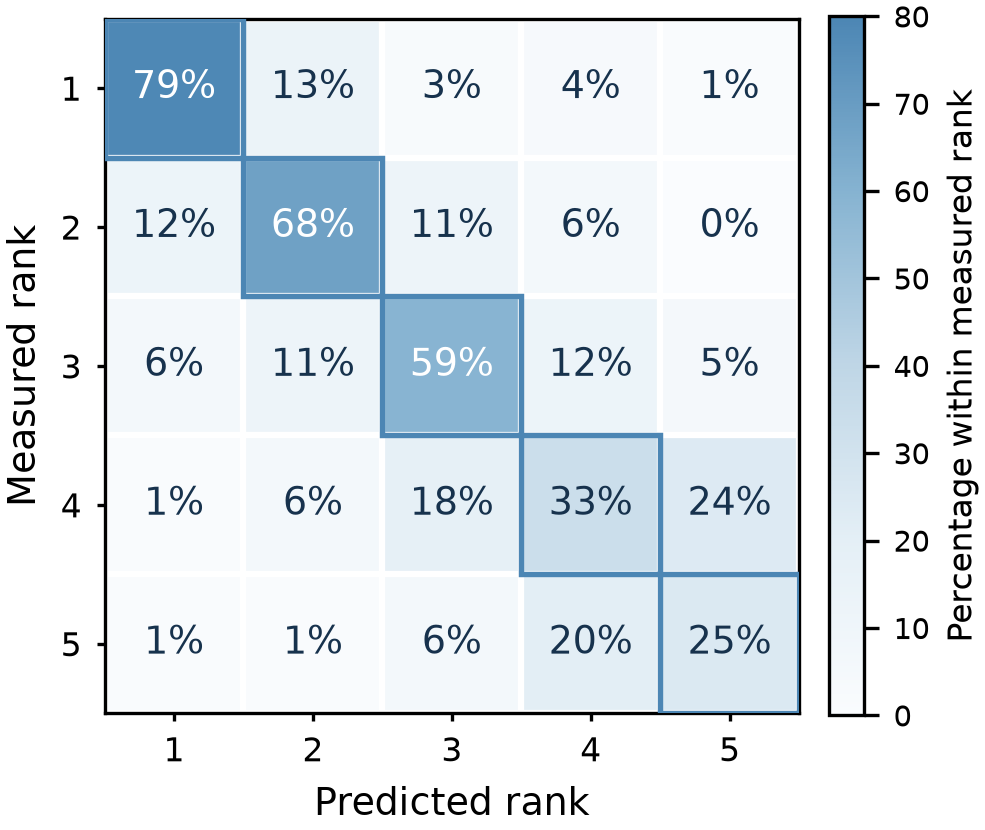}
\caption{Measured versus predicted ranks for the five leading elements.
Cells show specimen counts and colour indicates row percentage.}
\label{fig:rank-heatmap}
\end{wrapfigure}

\begin{figure}[t]
\centering
\captionsetup{font=scriptsize}
\includegraphics[width=\linewidth]{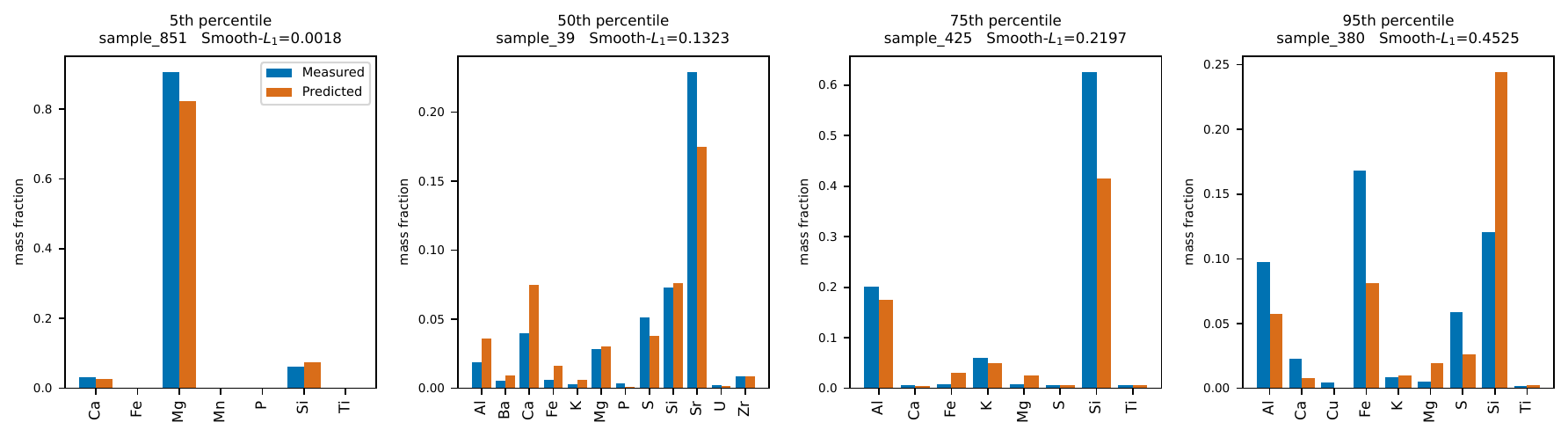}
\caption{Measured and predicted compositions at four percentiles of
specimen-level Smooth-$L_1$ error. Values are shown in mass-fraction units.} \label{fig:smoothl1-percentiles}
\end{figure}

\paragraph{How accurate is our prediction?} We switch gears and focus on the prediction produced by our model and discuss its distance to the true XRF targets. In Fig.~\ref{fig:smoothl1-percentiles} we show the specimen's true and predicted composition, for the 5th, 50th, 75th and 95th percentiles; these are ranked based on Smooth-$L_1$ error. In the 5th, the $\hat{y}_s$ closely follows the $y_s$ dominant and secondary elements. The median specimen retains all constituents, and, encouragingly, achieves accurate predictions. At the 95th percentile, the model still identifies all elements present in the specimen, but their relative magnitudes are less accurately recovered. 
The main limitation for difficult specimens therefore lies in estimating elemental proportions rather than detecting their presence.

Fig.~\ref{fig:rank-heatmap} depicts a variant of a confusion matrix, which shows the number of specimens where the rank between $y_s$ and $\hat{y}_s$ was preserved, in the first five positions of the composition vector. Ideally, this matrix should contain the value 114 in its diagonal; it would suggest perfect ranking. Interestingly, the first row suggests that out of 114 test specimens, we predicted the same dominant element 90 times, or predicted the correct dominant element as second. In fact, looking at the second (and the third row), we can realize that, typically, when an element was measured in the second (and the third) position, we mostly predict correctly the position, or we are off by a position, over or under but not much less. Overall, the model preserves the ordering of the most important elements (Fig.~\ref{fig:rank-heatmap}): the dominant element is recovered exactly in 79\% of specimens and falls within the predicted top three in 95\%, which is important for identifying a specimen's principal constituents.

%% file: sections/07_conclusion.tex
\section{Conclusion and limitations}

We show that elemental composition can be predicted accurately from hyperspectral imagery, given enough labeled samples. Framing the task as multiple-instance regression lets one label-free spectral encoder support a compact supervised head, which outperforms mean-spectrum models at matched downstream trainable capacity and approaches full fine-tuning.

% The study's main limitation is the range of specimens it can estimate limited to surface SWIR measurements of specimens, so generalization across instruments, acquisition conditions, and geological regions remains untested. The mineral-first references use different supervision, and the simplex threshold was selected using full-dataset statistics; they therefore represent conventional pipelines rather than strictly held-out, equal-information baselines. Spatially heterogeneous surfaces may also differ from averaged XRF measurements, while sparse elements limit reliable element-wise assessment. Predictions should complement rather than replace certified assays. 

The main limitation of this study is the range of specimens represented in the dataset. The proposed models perform well on the held-out specimens, but their ability to generalize to specimens with different compositions, geological origins, or acquisition conditions remains unknown. Evaluating the models on a broader and more diverse collection of specimens, ideally acquired with different instruments and under different conditions, is an important direction for future work.

% Cross-instrument transfer and uncertainty-aware evaluation are important next steps.